%% file: neurips_2023.tex
\documentclass{article}

    \PassOptionsToPackage{numbers, compress}{natbib}

\usepackage[final]{neurips_2023}

\usepackage[utf8]{inputenc} 
\usepackage[T1]{fontenc}    
\usepackage{hyperref}       
\usepackage{url}            
\usepackage{booktabs}       
\usepackage{amsfonts}       
\usepackage{nicefrac}       
\usepackage{microtype}      
\usepackage{xcolor}         
\usepackage{siunitx}
\usepackage{graphicx}
\usepackage{bbm}
\usepackage{pifont}

\usepackage{caption}
\usepackage{subfigure}
\graphicspath{ {./figures/} }

\input{math_commands.tex}

\title{SAMCLR: Contrastive pre-training on complex scenes using SAM for view sampling}

\author{Benjamin Missaoui \thanks{Corresponding author}\\ benjamin.missaoui@gmail.com \\ Georgia Institute of Technology
\And
Chongbin Yuan \\ chongbinyuan@gmail.com \\ National University of Singapore}

\begin{document}

\maketitle

\begin{abstract}

In Computer Vision, self-supervised contrastive learning enforces similar representations between different views of the same image. The pre-training is most often performed on image classification datasets, like ImageNet, where images mainly contain a single class of objects. However, when dealing with complex scenes with multiple items, it becomes very unlikely for several views of the same image to represent the same object category. In this setting, we propose SAMCLR, an add-on to SimCLR which uses SAM to segment the image into semantic regions, then sample the two views from the same region. Preliminary results show empirically that when pre-training on Cityscapes and ADE20K, then evaluating on classification on CIFAR-10, STL10 and ImageNette, SAMCLR performs at least on par with, and most often significantly outperforms not only SimCLR, but also  DINO and MoCo.
\end{abstract}

\section{Introduction}

Self-Supervised Learning (SSL) is now a well-established and reliable way to train neural networks to produce robust image representations, without ever providing labels. By first pre-training the model with a pretext task such as contrastive learning \citep{chen2020simple, he2019moco, caron2021emerging}, clustering \citep{caron2018deep, caron2020unsupervised}, pseudo-labeling \citep{grill2020bootstrap, chen2020exploring} or more recently masked image modeling \citep{MaskedAutoencoders2021, xie2021simmim}, SSL approaches have been able to produce features that compete or even outperform their supervised counterparts in many downstream tasks like image classification, object detection or image segmentation. This paper focuses on contrastive learning.

\paragraph{Motivation} Pre-training with a contrastive learning objective is most often performed on an image classification dataset, and ImageNet \citep{5206848} is the de-facto choice of the research community. While ImageNet is indeed a high-quality, large-scale and highly curated dataset, its popularity in contrastive learning also comes from the fact that it is an object-centric (in opposition with scene-centric) dataset. In the contrastive learning scheme, models are trained to output representations that are invariant to various image augmentations. If an image only contains one main object, it is highly likely that two random views from this image will also contain at least part of this object. However, contrastive pre-training on scene-centric datasets is known to yield models with less discriminative power, which ultimately underperform on downstream classification tasks \citep{10.5555/3495724.3496011}.

\begin{figure}[t]
\begin{center}
\includegraphics[width=11cm]{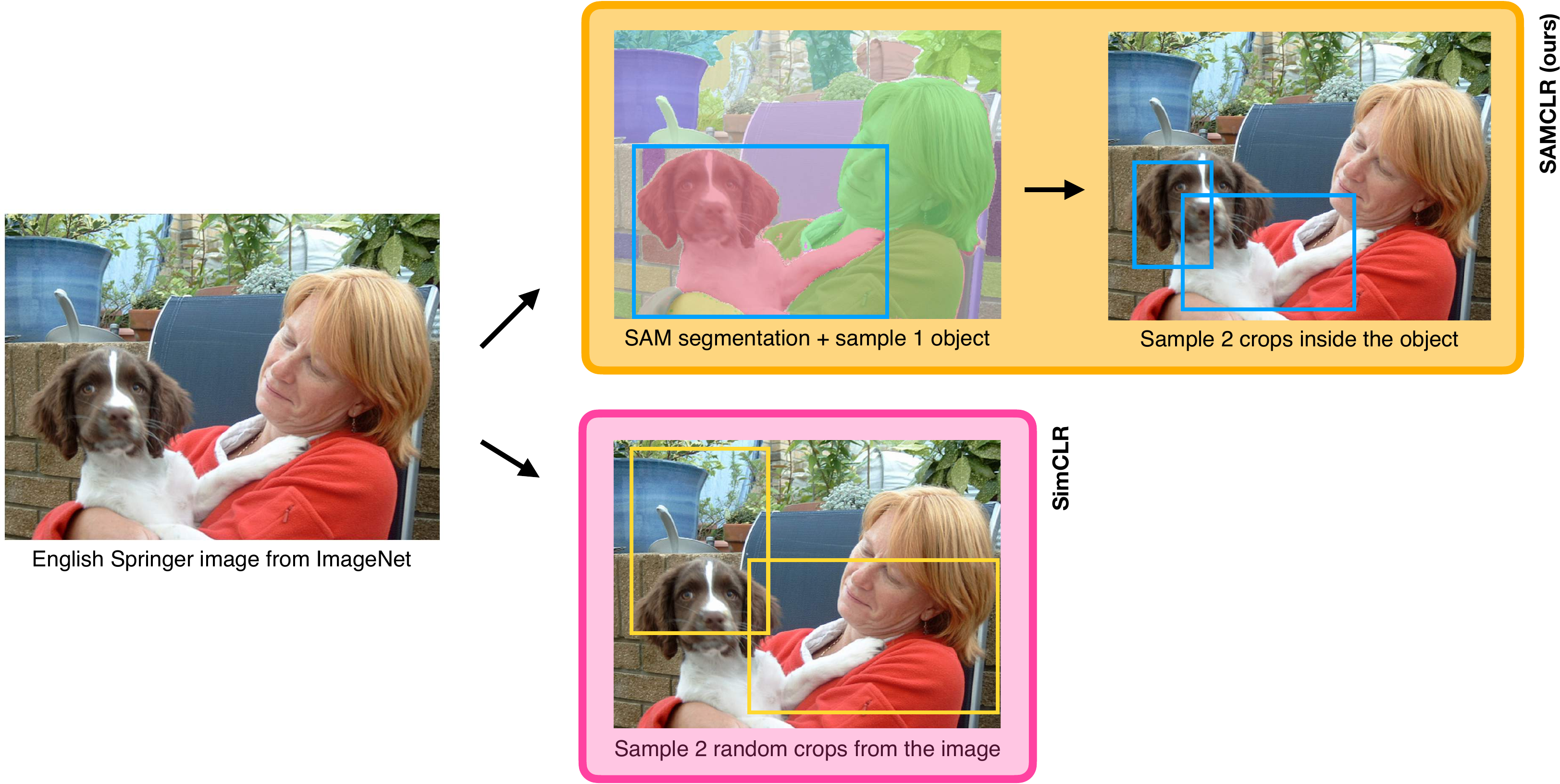}
\end{center}
\caption{View sampling process for SimCLR (bottom) and SAMCLR (ours, top). Using SAM to segment out the image avoids sampling views which represent different object categories, which in turn helps denoising the contrastive learning procedure.}
\label{fig:sam-ontology}
\end{figure}

Meanwhile, several foundation models for image segmentation have recently emerged, like SAM \citep{kirillov2023segany} and SEEM \citep{zou2023segment}, which both enable to divide any image into meaningful semantic regions. Our main idea is to use these segmentation models to perform object-level contrastive learning regardless of the the type of dataset used for pre-training. For a given scene, we propose to first use SAM to identify the different regions of interest in the image, then select one of those regions and sample the views inside of it, rather than the whole image (see Figure \ref{fig:sam-ontology}). This simple trick enforces that the different views are part of the same object. This idea derives from SimCLR \citep{chen2020simple}, which repels the views coming from different images. By doing so, SimCLR gives every image a different label, and since it is typically trained on ImageNet, it can be seen as giving a different label to each object. Thus, we aim at generalizing the SimCLR framework to complex scenes, by performing object-level contrastive learning regardless of the number of objects originally present in the scenes.

\paragraph{Contribution} Based on these observations, we propose \textbf{SAMCLR} (\textbf{SAM} + Sim\textbf{CLR}), a variation of SimCLR which first leverages SAM to segment the input image into semantic regions, then samples the views from one of these regions. By doing so, SAMCLR ensures that different views from the same image represent the same object, which helps denoising the learning process on complex scenes with multiple objects. When pre-training on datasets like Cityscapes \citep{Cordts2016Cityscapes} and ADE20K \citep{zhou2019semantic}, and evaluating the features on image classification on CIFAR-10 \citep{Krizhevsky2009LearningML}, STL10 \citep{Coates2011AnAO} and ImageNette \citep{imagenette}, SAMCLR performs at least on par with, and most often significantly outperforms not only SimCLR, but also other SSL baselines, like DINO \citep{caron2021emerging} and MoCo \citep{he2019moco}. 

\section{Method}

In this section, we first briefly review SAM and SAMCLR, and then detail our method which uses SAM to sample the views for SimCLR to learn from.

\subsection{Preliminaries}
\label{sec:preliminaries}

\paragraph{SAM}

Let $\textbf{\textsc{x}} \in \mathbb{R}^{H \times W \times C}$ be an image where $H, W$ and $C$ are the height, width and number of channels respectively. SAM first passes $\textbf{\textsc{x}}$ though a Vision Transformer encoder \citep{dosovitskiy2021image} $f_\theta$ in order to get an image embedding $z_{img} = f_\theta(\textbf{\textsc{x}})$. Then, a prompt encoder embeds the input prompts (clicks, box, text, mask...), we denote $z_{pr}$ this embedding. Finally, a transformer decoder $g_{\theta'}$ predicts a binary segmentation mask  $\hat{y} = g_{\theta'}(z_{img}, z_{pr})$, with $\hat{y} \in \mathbb{R}^{H \times W}$. When prompted with a regular grid of points, SAM will output a list of binary masks denoting different regions of the input image $\textbf{\textsc{x}}$. However, these regions are \textit{not} mutually exclusive and in most cases a few coarse regions will encompass all of the others. In our case, we are only interested in having a rough idea of the different elements of the scene. Thus, we only focus on the coarsest regions and we drop the others as they represent higher level of granularity. Then, we randomly select one of those regions (to get one object), and sample 2 views from it. 


\paragraph{SimCLR}



SimCLR attempts to enforce consistent representations between two augmentations from the same image, and does so by performing contrastive learning between images from the same batch. Given a batch of $n$ images $X = \{\textsc{x}^{(1)}, \dots,\textsc{x}^{(n)}\}$, SimCLR applies 2 random cropping and color jittering operations to each image. We denote $\Tilde{\textsc{x}}^{(i)}_1$ and $\Tilde{\textsc{x}}^{(i)}_2$ the views from image $\textsc{x}^{(i)}$ and $X_{aug}$ the resulting augmented batch $X_{aug} = \{\Tilde{\textsc{x}}^{(1)}_1$, $\Tilde{\textsc{x}}^{(1)}_2, \dots,  \Tilde{\textsc{x}}^{(n)}_1$, $\Tilde{\textsc{x}}^{(n)}_2 \}$, with $|X_{aug}| = 2n$. Then, a feature extractor $f_\theta$ and a MLP head $g_{\theta'}$ project each view to a feature vector $z^{(i)}_k = g_{\theta'}(f_\theta(\textsc{x}^{(i)}_k))$. Finally, the NT-Xent loss \citep{10.5555/3157096.3157304} brings closer the feature vectors of views from the same image, and repels the representations from the other images. Equation \ref{eq:ntxent} gives the loss for a pair of positive embeddings $z_i, z_j$.

\begin{equation}
    \ell_{i,j} = -\log \frac{\exp(\text{sim}(z_i, z_j)/\tau)}{\sum_{k=1}^{2n} \mathbbm{1}_{[k \neq i]}\exp(\text{sim}(z_i, z_j)/\tau)} \text{, with sim} (u, v) = \frac{u^Tv}{||u||\cdot ||v||}  
    \label{eq:ntxent}
\end{equation}




\subsection{SAMCLR: Instance-level contrastive learning}
\label{sec:region-level-ssl}

Our idea is very simple and is explained in Figure \ref{fig:sam-ontology}. Contrarily to SimCLR which samples 2 random views anywhere in the image, we first segment the image with SAM, then select randomly one of the semantic regions, and sample 2 views from it. This ensures that the two views represent the same object. Then, as in SimCLR, these 2 views undergo color jittering before being fed to the feature extractor, and before computing the distance to the embeddings of the other views from the batch. 

However, the randomness in SimCLR's view sampling process has the advantage of producing very diverse views. Limiting the sampling to views inside of objects would prevent our method from picking crops near object borders, which hinders variety. Additionally, padding around objects provides context that could help understanding. Thus, before selecting one object region, we multiply the height and width of each region produced by SAM by a constant factor $c$. We set $c=1.3$ for the rest of the paper.




\section{Experiments}

In this section, we evaluate the effectiveness of our method by pre-training SAMCLR, as well as three other SSL baseines (SimCLR, DINO and MoCo) on either CityScapes or ADE20K, then evaluating the resulting model on three classification datasets: CIFAR-10, STL10 and ImageNette. For the downstream classification task, we adopt two standard testing protocols, i.e. linear probing and KNN classification.

\paragraph{Implementation details} \label{par:implem} The pre-training of all SSL baselines is done with a ResNet18 \citep{7780459} with batch size 256. We use the implementations of SimCLR, DINO and MoCo available in Lightly \citep{susmelj2020lightly}, with default optimizers and hyperparameters. Our SAMCLR's implementation follows the one from SimCLR, and incorporates our view sampling procedure (see Sec. \ref{sec:region-level-ssl}). In practice, we perform the segmentation with HQ-SAM \citep{sam_hq}, a variant of SAM which produces slightly more precise and accurate masks. In order to speed up the view sampling process, we pre-compute and store the segmentation masks for all images of all datasets which will be used for pre-training. We also drop all regions with an area smaller than 1000px to avoid sampling views that would be too small. The temperature for SAMCLR is set to $\tau = 0.07$ as in MoCo. For SimCLR, SAMCLR and MoCo, the views are all resized to 128x128px. For DINO, the 2 global crops and 6 local crops are resized to 128x128px and 64x64px respectively. We run all experiments using a single RTX 4090 GPU with 24G memory. 
 
\paragraph{Pre-training on Cityscapes} We start with Cityscapes, which  contains images taken on the roads of 50 different cities. The images usually contain a lot of information, like other vehicles, pedestrians, traffic signs or buildings. We pre-train all SSL methods for 1600 epochs, using the \num[group-separator={,}]{2975} training images and we evaluate on image classification.

Table \ref{tab:cityscapes-res} sums up the results. We first note that the scores for all models are low overall. We believe this to be due to two main reasons. First, and as suggested before, contrastive pre-training is expected to fail on datasets which contain complex images (with more than 1 object), as views from the same image are no longer guaranteed to represent the same subject. On the other hand, the image distribution in Cityscapes is vastly different from the datasets used for evaluation. For example, ImageNette contains images of golf balls, parachutes and French horns, which are more than unlikely to appear in Cityscapes. Nevertheless, SAMCLR systematically improves over SimCLR by large margins (+21.9, +15.2 and +24.8 points on CIFAR-10, STL10 and ImageNette linear probing accuracy respectively). 

\begin{table}
\caption{KNN Top-1 accuracy and Linear probing for various SSL methods when pre-trained on Cityscapes and evaluated on image classification on CIFAR-10, STL10 and ImageNette. In this setting, SAMCLR systematically and significantly improves SimCLR and the other baselines.}
\label{tab:cityscapes-res}
\centering
\begin{tabular}{cccccccc}

\hline
&\multicolumn{3}{c}{KNN}    
& &                                  
\multicolumn{3}{c}{Linear}  \\ 
& CIFAR-10     & STL10     & ImageNette &  & CIFAR-10     & STL10     & ImageNette  \\
\hline

MoCo   &         44.0  &         41.2  &         45.4  &
       &         43.2  & 40.1 & 46.8 \\
DINO   &         38.8  &         36.7  &         40.8  &
       &         37.1  & 34.7 & 40.4 \\
SimCLR &         50.5  &         46.9  &         50.5  &
       &         42.5  & 40.8 & 42.7 \\
SAMCLR (ours) & \textbf{63.9} & \textbf{57.2} & \textbf{66.3} &
       & \textbf{63.4} & \textbf{56.0} & \textbf{67.5} \\
\hline
\end{tabular}
\end{table}

\paragraph{Pre-training on ADE20K} We then follow the same procedure to pre-train on ADE20K, which contains \num[group-separator={,}]{20210} training scenes from everyday life. We set the number of epochs to 250. The results are shown in Table \ref{tab:ade20k-res}. Once again, SAMCLR brings significant performance improvements over SimCLR (+8.6, +5.3 and +4.3 points on CIFAR-10, STL10 and ImageNette linear probing accuracy respectively). These gains make it outperform MoCo on CIFAR-10, and close the gap on the other benchmarks. We note that this time, all models perform better compared to the pretraining on Cityscapes. This is easily explained by ADE20K having a higher volume of images (20K compared to 3K), and a distribution closer to the downstream tasks. For example, ADE20K contains airplane images, which are very unlikely to appear in Cityscapes whereas \textit{airplane} is actually one of the class labels for both CIFAR-10 and STL10.

\begin{table}
\caption{KNN Top-1 accuracy and Linear probing for various SSL methods when pre-trained on ADE20K and evaluated on image classification on CIFAR-10, STL10 and ImageNette. SAMCLR systematically improves SimCLR but looses to MoCo on STL10 and ImageNette.}
\label{tab:ade20k-res}
\centering
\begin{tabular}{cccccccc}

\hline
&\multicolumn{3}{c}{KNN}    
& &                                  
\multicolumn{3}{c}{Linear}  \\ 
& CIFAR-10     & STL10     & ImageNette &  & CIFAR-10     & STL10     & ImageNette \\
\hline

MoCo   &         63.1  &\textbf{61.9}  & \textbf{71.8}  &
       &         60.7  & \textbf{60.7} & \textbf{73.5} \\
       
DINO   &         52.4  &         55.8  &         65.4  &
       &         48.9  & 51.7 & 62.8 \\

SimCLR &         60.6  &         59.1  &         68.8  &
       &         57.0  & 54.5 & 66.5 \\

SAMCLR (ours) & \textbf{65.5} & 59.3   &         69.1 &
       & \textbf{65.6} & 59.8 & 70.8 \\
\hline
\end{tabular}
\end{table}

\section{Future work}

This paper presents preliminary results on the use of SAM (and foundation models more broadly) to support the pretraining of other unsupervised methods. While we introduced SAMCLR as an add-on for SimCLR, it would be interesting to see if other contrastive learning approaches also benefit from it. Since SAMCLR only affects the view sampling process, it would be easily to integrates it into other baselines, especially MoCo which is very similar to SimCLR. Furthermore, pre-training SSL approaches on scene-centric datasets has shown to yield better feature extractors for downstream dense prediction tasks, like object detection or image segmentation \citep{10.5555/3495724.3496011}. It would be interesting to see if SAMCLR follows this trend and outperforms the other baselines in such benchmarks.

\section{Conclusion}

In this paper, we introduced a simple view sampling strategy, coined SAMCLR, that can be plugged into SimCLR to improve its learning capability on datasets with complex scenes. SAMCLR relies on SAM to ensure coherence and consistency between the views sampled from the same image. We validated the effectiveness of our method through experiments on Cityscapes and ADE20K, two datasets originally designed for segmentation tasks. We showed empirically that in these settings, SAMCLR systematically improves over SimCLR, and most often by a significant margin. 


\bibliography{neurips_2023}
\bibliographystyle{plainnat}

\newpage
\appendix
\section{Supplementary Material}


\subsection{Pre-training on an object-centric dataset (ImageNette)}

While we aim at designing a SSL approach that enables pre-training on scene-centric datasets, it remains interesting to see how SAMCLR performs when pre-trained on an image classification dataset. Thus, we pre-train and evaluate both SAMCLR and the other SSL baselines on ImageNette with the same experimental setting as in Section \ref{par:implem}. We set the number of epochs to 800. Table \ref{tab:imagenette-res} shows the results. We notice that SAMCLR underperforms in this setting. With an image classification dataset, we believe that our view sampling method hinders the variety of possible views fed to the contrastive loss, preventing the model from generalizing.

\begin{table}[h]
\caption{KNN Top-1 accuracy and Linear probing for various SSL baselines when pre-trained and evaluated on ImageNette.}
\label{tab:imagenette-res}
\centering
\begin{tabular}{ccc}

\hline
& KNN & Linear \\
\hline

SwaV          & \textbf{89.5} & \textbf{91.9} \\       
DINO          &         85.9  &         88.6  \\
SimCLR        &         89.0  &         88.5  \\
SAMCLR        &         85.1  &         85.8  \\
\hline
\end{tabular}
\end{table}

\subsection{SimCLR vs SAMCLR - Learning curves}

\begin{figure}[h]
    \centering
    \subfigure[]{\includegraphics[width=0.32\textwidth]{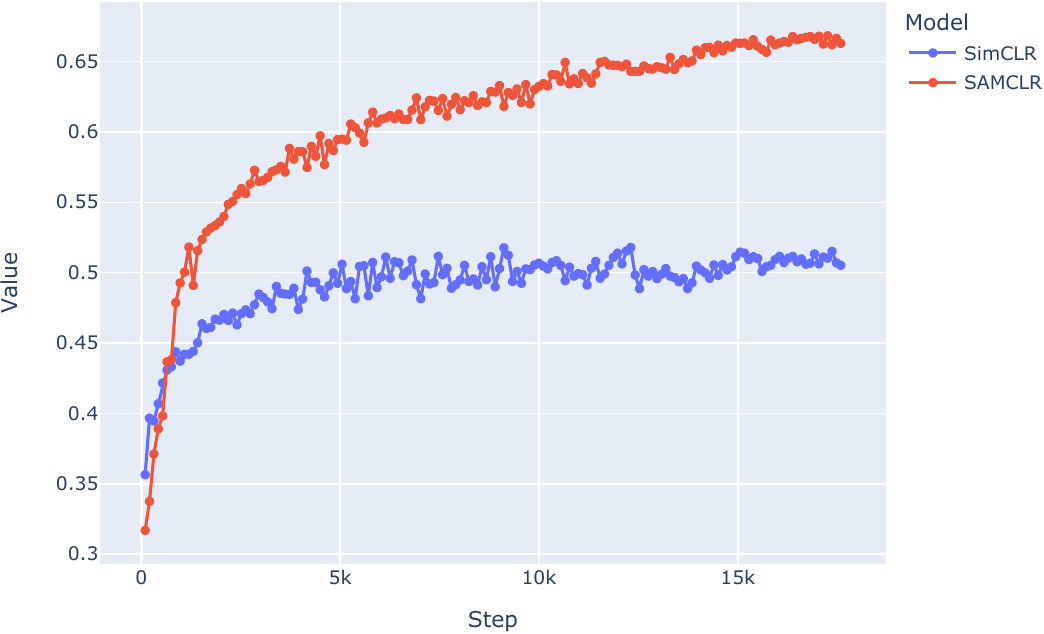}} 
    \subfigure[]{\includegraphics[width=0.32\textwidth]{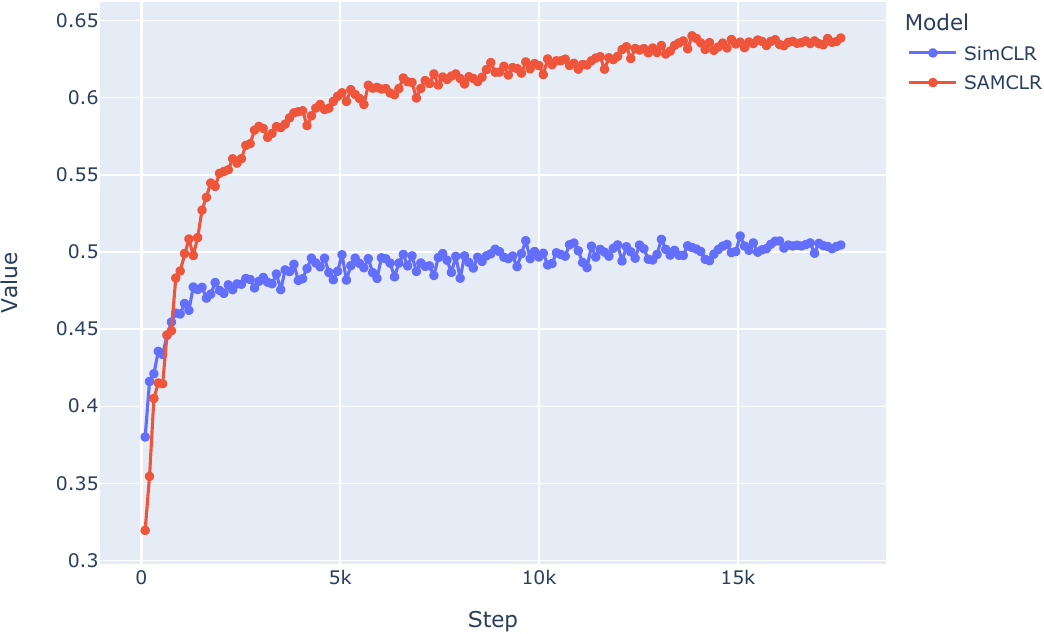}} 
    \subfigure[]{\includegraphics[width=0.32\textwidth]{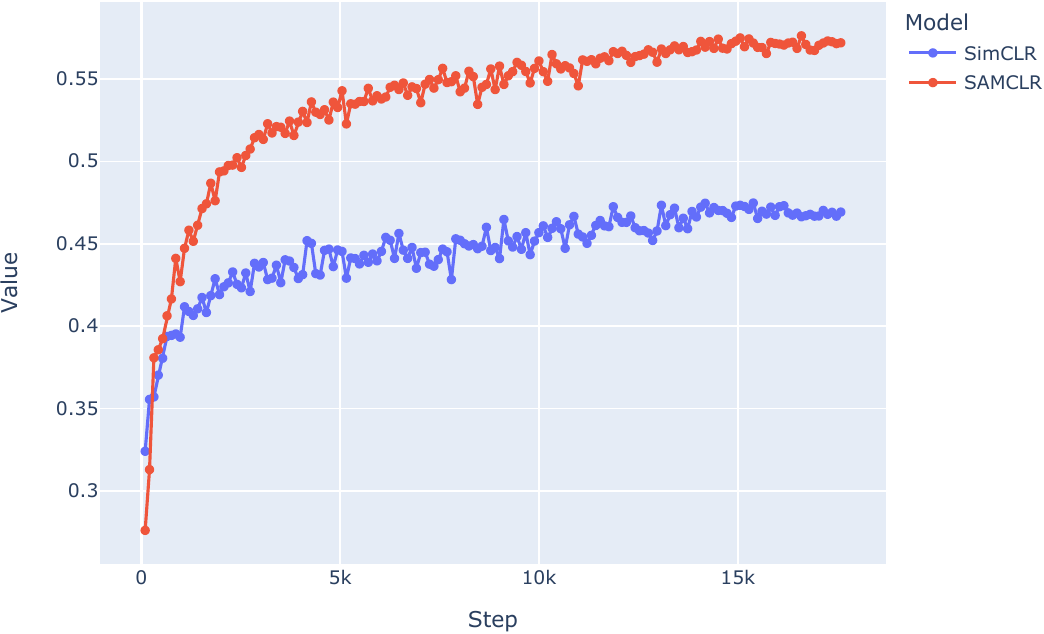}}
    \caption{KNN accuracy vs optimization steps for SimCLR (blue) and SAMCLR (ours, red), pre-trained on Cityscapes and evaluated on ImageNette (a), CIFAR-10 (b) and STL10 (c).}
    \label{fig:simclr-vs-samclr}
\end{figure}

To give a better idea of how SAMCLR fares against SimCLR and gauge the influence of our view sampling process, we display in Figure \ref{fig:simclr-vs-samclr} the evolution of KNN classification accuracy as a function of the number of steps for both SAMCLR and SimCLR. Both models are pretrained on Cityscapes. The corresponding final results are given in Table \ref{tab:cityscapes-res}. SAMCLR basically outperforms SimCLR throughout the whole learning process. 

We also note that the accuracy for SAMCLR is still seemingly going up at the end of training (1600 epochs). We thus attempt to train it for longer (3200 epochs) and observe an about 1.5 point increase on ImageNette KNN accuracy. We explain SAMCLR's appetite for longer training by the fact that, for each image, we sample the views from one object. Thus, over one epoch, the model will only see a single object of each image. It will therefore take longer for the model to see all objects of all images. SimCLR, on the other end, samples the views randomly, and thus has more chances to sample views with several objects, and it may even sample large crops that would reveal a large part of the original image.

\subsection{Related works}

\paragraph{Self-supervised learning of image features.} 

\label{sec:related-works}

Many different approaches have been proposed for learning good representations from an unlabeled set of images. This paper draws inspiration from the contrastive learning literature \citep{chen2020simple, he2019moco, caron2021emerging, grill2020bootstrap, chen2020exploring}, which enforces similar representations between two or more views of the same image. Every method brings its own novelty, like BYOL \cite{grill2020bootstrap} which does not use negative samples. Other approaches include clustering, like \citep{caron2018deep, caron2020unsupervised}. In particular, SwaV \citep{caron2020unsupervised} performs online clustering, making it one of the few clustering methods to be easily trainable at scale. 

\paragraph{Sample selection and choice of dataset} Recent work \cite{10.5555/3495724.3496011} shows that the performance of current SSL schemes heavily depends on the dataset used for pretraining. For example, a MoCo \citep{he2019moco} trained on MSCOCO leads to less discriminative power and lower classification accuracy than when pretrained on MSCOCO Boxes (a dataset they obtain by cropping the annotated bounding boxes from COCO). This observation motivated the current paper. While some other recent work has been tackling the view sampling process in contrastive learning  \citep{DBLP:journals/corr/abs-2011-11765, 10208400}, we believe that the recent development of foundational segmentation models opens up new possibilities in the field.

\end{document}

%% file: math_commands.tex
\usepackage{amsmath,amsfonts,bm}

\def\eqref#1{equation~\ref{#1}}

\def\1{\bm{1}}

\DeclareMathAlphabet{\mathsfit}{\encodingdefault}{\sfdefault}{m}{sl}
\SetMathAlphabet{\mathsfit}{bold}{\encodingdefault}{\sfdefault}{bx}{n}

